\documentclass[letterpaper, 10 pt, conference]{ieeeconf}
\IEEEoverridecommandlockouts 
\usepackage{tikz}
\newcommand\copyrighttext{%
  \footnotesize © 2022 IEEE.  Personal use of this material is permitted.  Permission from IEEE must be obtained for all other uses, in any current or future media, including reprinting/republishing this material for advertising or promotional purposes, creating new collective works, for resale or redistribution to servers or lists, or reuse of any copyrighted component of this work in other works.}
\newcommand\copyrightnotice{%
\begin{tikzpicture}[remember picture,overlay]
\node[anchor=south,yshift=10pt] at (current page.south) {\fbox{\parbox{\dimexpr\textwidth-\fboxsep-\fboxrule\relax}{\copyrighttext}}};
\end{tikzpicture}%
}

\title{\LARGE \bf
Lightweight Monocular Depth Estimation through Guided Decoding}

\author{Michael Rudolph$^{1}$, Youssef Dawoud$^{2}$, Ronja Güldenring$^{3}$, Lazaros Nalpantidis$^{3}$ and Vasileios Belagiannis$^{2}$ 
\thanks{$^{1}$University of Duisburg-Essen, Essen, Germany,
        {\tt\small michael.rudolph@uni-due.de}. Work done at Ulm University.}%
\thanks{$^{2}$Ulm University, Ulm, Germany,
        {\tt\small $\{$first.last$\}$@uni-ulm.de}}%
\thanks{$^{3}$DTU -- Technical University of Denmark, Kgs. Lyngby, Denmark,
        {\tt\small \{ronjag, lanalpa\}@elektro.dtu.dk}%
}}

\newcommand{\subparagraph}{}
\usepackage{graphicx}
\usepackage{amsmath}
\usepackage{amssymb}
\usepackage{multirow}
\usepackage{subfig}
\usepackage[hyphens]{url}
\usepackage[strings]{underscore}
\begin{document}

\maketitle
\thispagestyle{empty}
\pagestyle{empty}
\copyrightnotice

%%%%%%%%%%%%%%%%%%%%%%%%%%%%%%%%%%%%%%%%%%%%%%%%%%%%%%%%%%%%%%%%%%%%%%%%%%%%%%%%
\begin{abstract}
We present a lightweight encoder-decoder architecture for monocular depth estimation, specifically designed for embedded platforms. Our main contribution is the Guided Upsampling Block (GUB) for building the decoder of our model. Motivated by the concept of guided image filtering, GUB relies on the image to guide the decoder on upsampling the feature representation and the depth map reconstruction, achieving high resolution results with fine-grained details. Based on multiple GUBs, our model outperforms the related methods on the NYU Depth V2 dataset in terms of accuracy while delivering up to 35.1 fps on the NVIDIA Jetson Nano and up to 144.5 fps on the NVIDIA Xavier NX. Similarly, on the KITTI dataset, inference is possible with up to 23.7 fps on the Jetson Nano and 102.9 fps on the Xavier NX. Our code and models are made publicly available\setcounter{footnote}{3}\footnote{https://github.com/mic-rud/GuidedDecoding}.

\end{abstract}

%%%%%%%%%%%%%%%%%%%%%%%%%%%%%%%%%%%%%%%%%%%%%%%%%%%%%%%%%%%%%%%%%%%%%%%%%%%%%%%%

\section{INTRODUCTION}
Depth estimation is a major perception component of robotics systems, which can be also combined with downstream vision tasks~\cite{Gueldenring2021_IROS, nissler2015omg, wiederer2020traffic}. For instance, self-localization~\cite{engel2019deeplocalization}, visual odometry~\cite{Kostavelis2016_IJARS} or object detection~\cite{Kovacs2021_ICVS} often rely on depth information in the context of automated driving or robot navigation. While RGB-D and stereo cameras, as well as laser and radar sensors provide accurate depth estimates, they can be costly and often only deliver sparse depth information. In contrast, monocular depth estimation is an inexpensive and easily deployable approach that has been recently well-developed thanks to deep learning advances.

\begin{figure}[!ht]
    \centering
    \includegraphics[width=.45\textwidth]{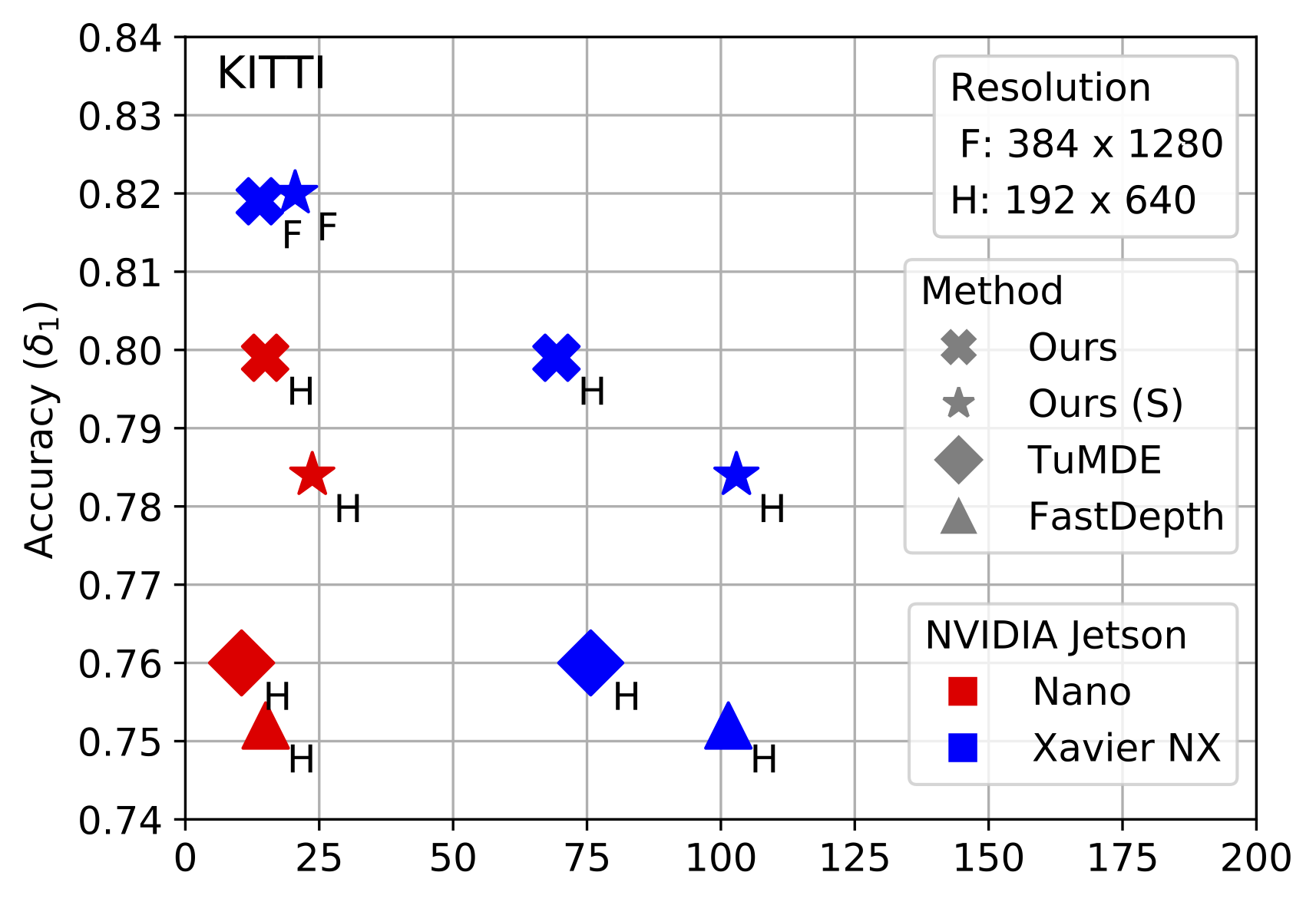}
    \includegraphics[width=.45\textwidth]{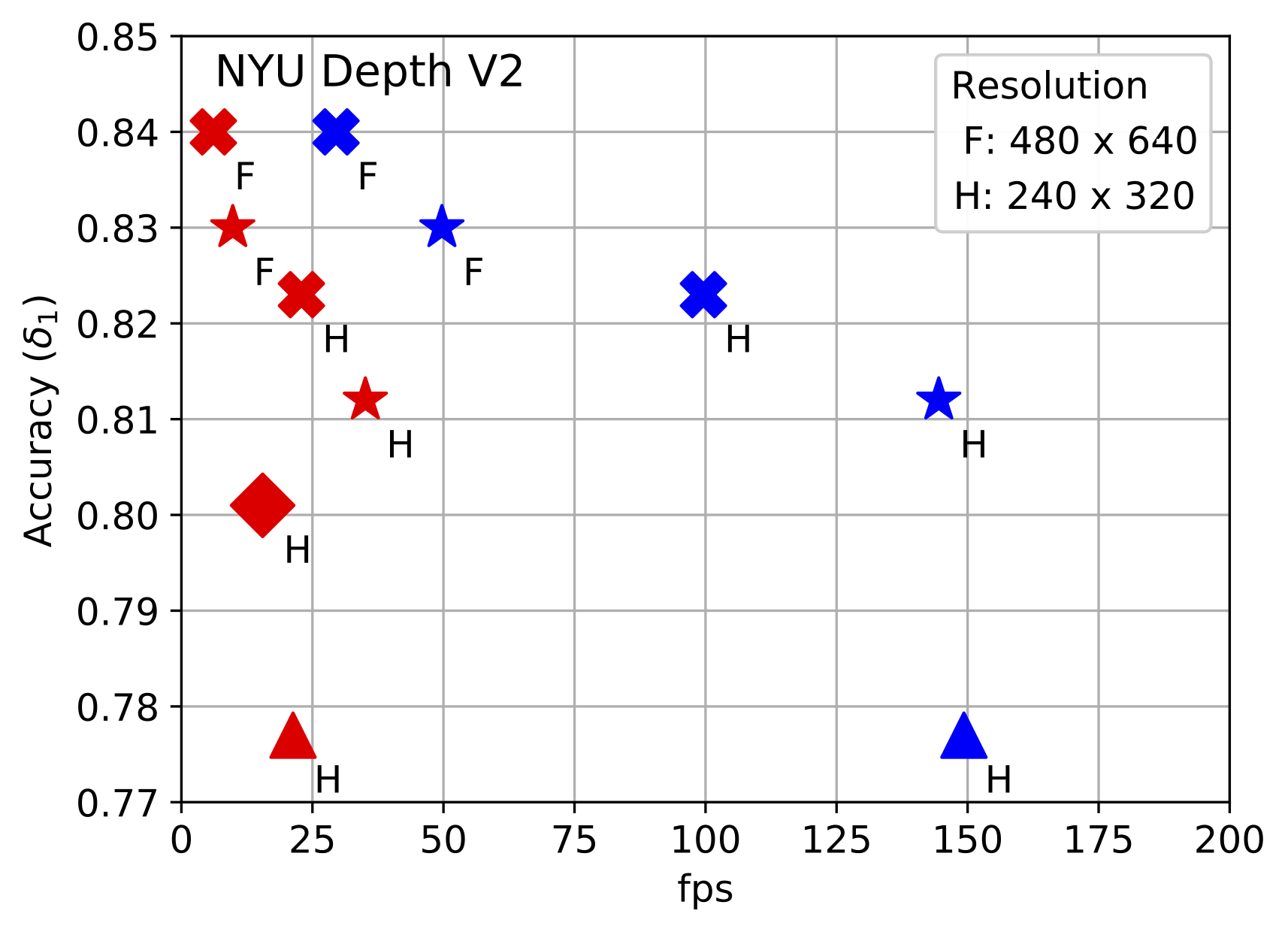}
    %\includegraphics[scale=.55, trim={0cm 0cm 0cm 0cm}]{images/KITTI_Trade-off.pdf}
    %\includegraphics[scale=.55, trim={0cm 0cm 0cm 0cm}]{images/NYU_Trade-off.pdf}
    %\vspace*{-2mm} 
    \caption{Our proposed architecture outperforms related monocular depth estimation methods wrt. accuracy on KITTI \cite{geiger_vision_2013} and NYU Depth V2 \cite{silberman_indoor_2012}, while delivering competitive inference speed on the NVIDIA Jetson Nano and the NVIDIA Jetson Xavier NX devices when comparing to FastDepth \cite{wofk_fastdepth_2019} and TuMDE \cite{tu_efficient_2021}, which were re-trained and evaluated using our described procedure. In addition, our extra light model (Ours-S) reaches solid performance at a high frame rate.}
    \label{fig:NYU_delta_fps}
    %\vspace*{-2mm}

\end{figure}

A plethora of monocular depth estimation approaches has been proposed based on convolutional neural networks \cite{laina_deeper_2016, alhashim_high_2018, lee_big_2019} and different types of supervision \cite{godard_unsupervised_2017, godard_digging_2019, garg_unsupervised_2016}. The majority of the existing approaches, though, targets accuracy over real-time performance \cite{bhat_adabins_2021, song_monocular_2021, ranftl_vision_2021}. Consequently, these approaches cannot deliver real-time execution on devices with constrained resources. 

Efficient approaches \cite{wofk_fastdepth_2019, tu_efficient_2021} therefore often employ hardware-specific compilation \cite{chen_tvm_2018} and model compression \cite{yang_netadapt_2018, he_amc_2018, belagiannis_adversarial_2018} to achieve higher throughput. Additionally, only low resolution inputs and outputs are considered, resulting in the loss of fine-grained details and blurred edges in the predicted depth maps. 

In this paper, we present a lightweight encoder-decoder architecture for monocular depth estimation. Our motivation is delivering real-time performance on embedded systems without the necessity to compress the model or rely on specific hardware compilation. To reconstruct high resolution depth maps, we propose the Guided Upsampling Block (GUB) for designing the decoder while relying on a standard encoder from the literature. Inspired by guided image filtering \cite{he_guided_2010, li_deep_2016}, the proposed GUB makes use of the input image at different resolutions to guide the decoder on upsampling the feature representation as well as for the final depth map reconstruction. By stacking multiple GUBs in sequence, we build a cost-efficient decoder, allowing to infer detailed depth maps on constrained hardware. Moreover, our evaluation shows that we reach a good balance between accuracy and speed when comparing with the related work on KITTI \cite{geiger_vision_2013} and NYU Depth V2 \cite{silberman_indoor_2012} datasets.

To sum up, we propose the Guided Upsampling Block (GUB) for building the decoder of a convolutional neural network for monocular depth estimation. Our approach outperforms the related methods that target embedded systems in terms of accuracy for both KITTI and NYU Depth V2 benchmarks while achieving inference speed suitable for real-time applications, as illustrated in Fig.~\ref{fig:NYU_delta_fps}.

\section{RELATED WORK}
\label{MonDepthRelated}
The problem of monocular depth estimation with deep neural networks has been well studied over the past few years. At first, Eigen \textit{et al.} \cite{eigen_depth_2014} proposed to use a coarse-scale network, relying on fully connected layers to achieve a global receptive field, in order to give a coarse depth prediction, which is then refined locally by a fine-scale network.
In contrast, by using a pre-trained ResNet \cite{he_deep_2016} model as feature extractor, Laina \textit{et al.} \cite{laina_deeper_2016} proposed a fully convolutional encoder-decoder architecture for depth estimation.
Based on this approach, recent methods heavily rely on convolutional neural networks by using pre-trained, general-purpose backbones as encoder and focusing on decoder design \cite{alhashim_high_2018, fu_deep_2018, lee_big_2019, song_monocular_2021}. \\ 
To increase the receptive field, Atrous Spatial Pyramid Pooling (ASPP) \cite{chen_deeplab_2018} modules from semantic segmentation are leveraged in recent architectures \cite{fu_deep_2018, lee_big_2019, song_monocular_2021} and
skip-connections from the encoder to the decoder are used to improve optimization \cite{alhashim_high_2018, lee_big_2019, song_monocular_2021}. 
Song \textit{et al.} \cite{song_monocular_2021} achieve state-of-the-art performance based on images from the Laplacian pyramid as guidance in the decoder.
More recently, Ranftl \textit{et al.} \cite{ranftl_vision_2021} show that self-attention based architectures like vision transformer \cite{dosovitskiy_image_2021} are capable of outperforming convolutional neural networks when provided with enough training data. Apart from architectural novelties, advances have been made to allow training with fewer data and computationally efficient augmentations \cite{alhashim_high_2018}, as well as using additional supervision through virtual normals \cite{yin_enforcing_2019}.\\
As the above-discussed approaches are generally too complex to deploy on embedded hardware, Wofk \textit{et al.} \cite{wofk_fastdepth_2019} proposed the usage of a pre-trained MobileNetV2 encoder in combination with a lightweight decoder leveraging depth-wise separable convolutions. Tu \textit{et al.}~\cite{tu_efficient_2021} rely on a similar architecture to allow deployment to the NVIDIA Jetson Nano, but further simplify the decoder. Poggi \textit{et al.}~\cite{poggi_towards_2018} proposed a hand-designed custom architecture to allow inference on the CPU of the Raspberry Pi3, while Aleotti \textit{et al.}~\cite{aleotti_real-time_2020} investigate the deployment of the model on mobile phones.
To increase throughput, input resizing is performed through center-cropping in \cite{poggi_towards_2018}, \cite{wofk_fastdepth_2019} and \cite{tu_efficient_2021}, which however results in evaluating on self-defined crops, and therefore inconsistencies when comparing performance.
To further increase inference speed, models are compiled to TVM \cite{chen_tvm_2018} in \cite{wofk_fastdepth_2019} and \cite{tu_efficient_2021}. Additionally, pruning with the NetAdapt \cite{yang_netadapt_2018} framework is used by Wofk \textit{et al.} \cite{wofk_fastdepth_2019} to reduce the model size while Tu \textit{et al.} \cite{tu_efficient_2021} reduce model size with the help of a reinforcement learning algorithm \cite{he_amc_2018}. 
While these post-processing steps greatly increase the throughput of lightweight models, they make deployment complex and time-consuming. We aim at closing this gap by proposing a method that achieves fast and precise inference while being easy to deploy by only using TensorRT \cite{nvidia_tensorrt_2021} for inference. To allow a fair comparison when performing inference on lower resolution, we bilinearly upscale the prediction to allow evaluation on the original evaluation crop of the respective dataset. 

It is worthwile to investigate efficient semantic segmentation methods, as these approaches also produce pixel-wise predictions, but often rely on bilinear interpolation to replace decoders \cite{yu_bisenet_2018, hong_deep_2021}. Simply applying these approaches to monocular depth estimation results in blurred depth maps of insufficient quality; this stresses the importance of the decoder for monocular depth estimation. Our decoder design is inspired by guided image filtering networks \cite{li_deep_2016} combined with the design principles of standard decoders. By using image guidance in the decoder, we are able to reconstruct fine-grained details in the prediction. Additionally, this allows to reduce the size of feature maps in the decoder to reduce computational complexity. By stacking multiple blocks sequentially, we can also reduce the size of convolutional kernels while maintaining a similar receptive field, further reducing complexity.

\section{METHOD}

%\subsection{Problem Definition}

Assume the availability of the training set $\mathcal{T} = \{(\mathbf{x}, \mathbf{y})_i \}_{i=1}^N$ where $\mathbf{x}_{i} \in \mathbb{R}^{H\times W \times 3}$ is an RGB image and $\mathbf{y}_{i} \in \mathbb{R}^{H\times W}$ is the corresponding real-valued ground-truth depth map with the same resolution as the image. Our objective is to utilize $\mathcal{T}$ for learning the image to depth mapping $f_\theta: \mathbb{R}^{H\times W \times 3} \rightarrow \mathbb{R}^{H \times W}$ with the deep neural network $f_\theta$, parametrized by $\theta$. Furthermore, the neural network $f_\theta$ is decomposed into the encoder $\psi$ and the decoder $\phi$ parts, defined as:
\begin{equation}\label{eq:encodeco}
    f_\theta(\mathbf{x}) = \psi(\phi(\mathbf{x})),
\end{equation}
where the encoder $\phi(\cdot)$ maps the input image to the latent space and the decoder $\psi(\cdot)$ reconstructs the depth map from it. In this work, our contribution is the decoder part of the network, while we rely on a standard architecture for the encoder. We propose the Guided Upsampling Block for the decoder to enable high-quality depth map reconstruction at a lower computational cost.

\begin{figure}[tp]
    \centering
    \includegraphics[scale=.37, trim={0cm 9cm 19cm -.4cm}, clip]{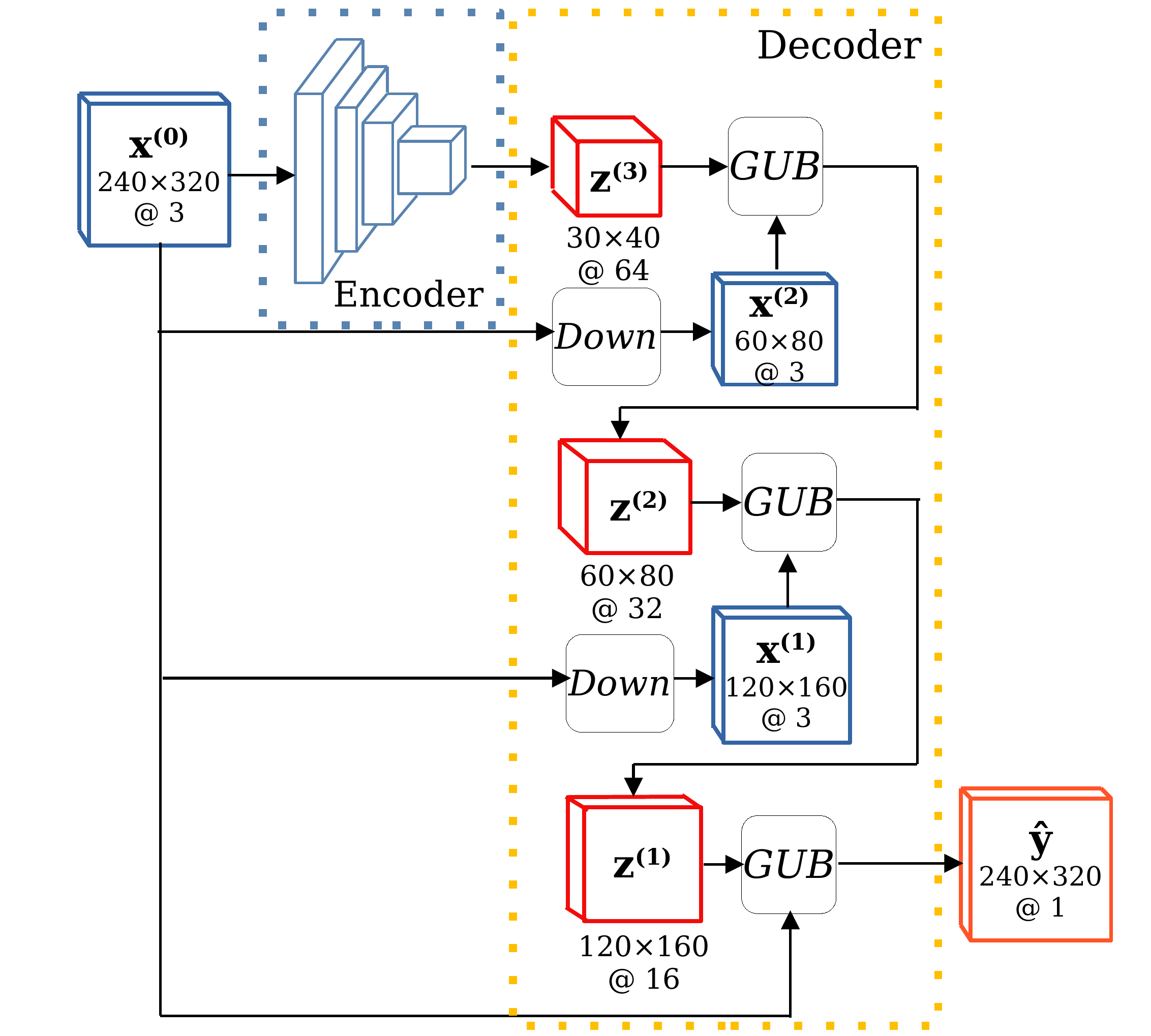}
    \caption{GuideDepth architecture, consisting of DDRNet-23-slim encoder and a novel decoder, consisting of three Guided Upsamling Blocks ($GUB$).  $Down$ denotes bilinear interpolation.}
    \label{fig:GuideDepth}
\end{figure}

\subsection{Guided Upsampling Block}
\label{sec:GUB}
Our Guided Upsampling Block (GUB) is motivated by the idea of guided image filtering \cite{he_guided_2010} and the more recent ideas of learning these filters with convolutional neural networks~\cite{li_deep_2016}. In guided image filtering, the usage of a guidance image helps to enhance the degraded target image where degradation can occur, for example, due to the low spatial resolution of the target image. In our context, we propose to use the input image of the model at different resolutions for guiding the feature upsampling in the decoder part (see Fig.~\ref{fig:GuideDepth}). A GUB takes as input the image and the decoder's feature representation to deliver the upsampled feature representation. We define it as
\begin{equation}
    \mathbf{z}^{(j)} = GUB(\mathbf{z}^{(j+1)}, \mathbf{x}^{(j)}),
\end{equation}
where $\mathbf{z}$ refers to feature maps and $\mathbf{x}$ refers to guidance images. Their spatial resolution is indexed by $j$, being $\frac{1}{2^j}$ of the input resolution. Moreover, we design the decoder with several GUBs to progressively upsample the feature representation of each decoder layer and finally the reconstructed depth map, as shown in Fig.~\ref{fig:GuideDepth}. 

\begin{figure}[tp]
    \centering
    \subfloat[Guided Upsampling Block $GUB$]{\includegraphics[trim={0cm 20cm 12.5cm 0cm}, clip, scale=.29]{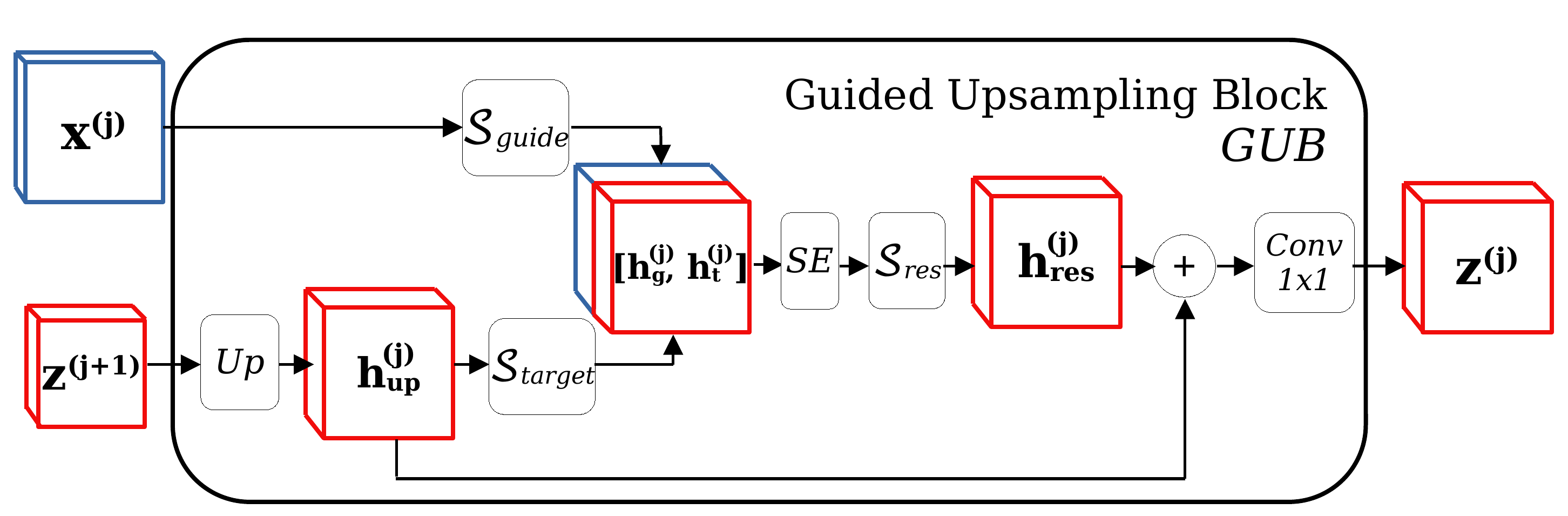}}\\
    \subfloat[Stacked Convolution $S$]{\includegraphics[trim={0cm 24cm 20cm .5cm}, clip, scale=.3]{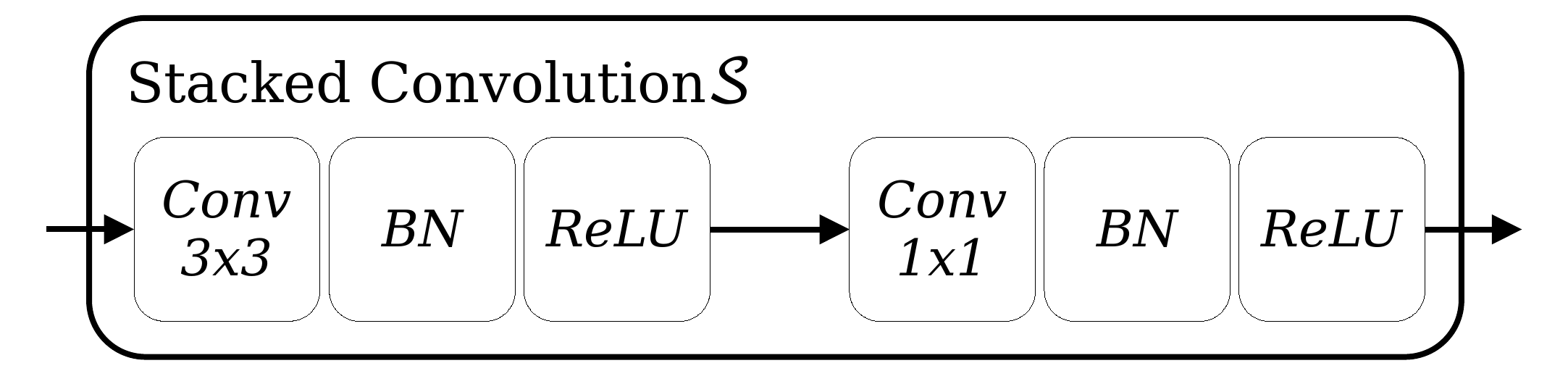}}
    \caption{The Guided Upsampling Block ($GUB$) in (a) uses the input image $\mathbf{x}^{(j)}$ as guidance to upsample the feature representation $\mathbf{z}^{(j+1)}$ by a factor of 2, resulting in $\mathbf{z}^{(j)}$. The fundamental block operation is the \textit{stacked convolution} $\mathcal{S}$, shown in (b). \textit{Up} denotes bilinear interpolation by factor 2; \textit{SE} represents the Squeeze-and-Excitation module \cite{hu_squeeze-and-excitation_2018}.}
    \label{fig:GUB}
\end{figure}

The GUB operations are illustrated in Fig.~\ref{fig:GUB}. A GUB upsamples a feature map by scale factor 2 while reconstructing fine-grained details from the guidance image. The fundamental block operation is the \textit{stacked convolution} $\mathcal{S}$ that consists of a convolution with kernel size 3, a batch normalization layer and a ReLU activation, followed by the same operations with a kernel size of 1 as illustrated in Fig.~\ref{fig:GUB} (b). We rely on the stacked convolution operation to collect three type of features. At first, it is used to extract features $\mathbf{h}_g^{(j)}$ from the guidance image $\mathbf{x}^{(j)}$, defined as:
\begin{equation}
    \mathbf{h}_g^{(j)} = \mathcal{S}_{guide}(\mathbf{x}^{(j)}),
\end{equation}
where $\mathcal{S}_{guide}$ denotes the stacked convolution. These features guide the upsampling step. Second, the feature representation of the encoder $\mathbf{z}^{(j+1)}$ is upsampled by a factor of 2 through bilinear interpolation $Up(\cdot)$. Next, another stacked convolution $\mathcal{S}_{target}$ is used to refine the upsampled representation. These operations are represented as:
\begin{align}
    \mathbf{h}_{up}^{(j)} &= Up(\mathbf{z}^{(j+1)}) \\
    \mathbf{h}_{t}^{(j)} &= \mathcal{S}_{target}(\mathbf{h}_{up}^{(j)}).
\end{align}
Finally, the feature tensors $\mathbf{h}_{g}^{(j)}$ and $\mathbf{h}_{t}^{(j)}$ are concatenated to create a joint representation, as shown in Fig.~\ref{fig:GUB}. To attend to the most important feature channels, we apply the Squeeze-and-Excitation module \cite{hu_squeeze-and-excitation_2018} $SE(\cdot)$, followed by the last stacked convolution operation $\mathcal{S}_{res}$. We describe these operations as:
\begin{equation}
\begin{aligned}
    \mathbf{h}_{res}^{(j)} = \mathcal{S}_{res}(SE([\mathbf{h}_{t}^{(j)}, \mathbf{h}_{g}^{(j)}])),
\end{aligned}
\end{equation}
where we consider $\mathbf{h}_{res}^{(j)}$ as the correction term for the upsampled feature representation $\mathbf{h}_{up}^{(j)}$. For that reason, we add $\mathbf{h}_{res}^{(j)}$ with $\mathbf{h}_{up}^{(j)}$ as we illustrate in Fig.~\ref{fig:GuideDepth} (a). At last, we apply a $1\times1$ convolution to reduce the number of feature maps. 

\subsection{Complete Encoder-Decoder}
As defined in Eq.~\ref{eq:encodeco}, our model consists of the encoder $\phi(\cdot)$ and the decoder $\psi (\cdot)$. As encoder, we choose the DDRNet-23-slim \cite{hong_deep_2021}, pre-trained on the ImageNet database~\cite{deng_imagenet_2009}. We pick up this semantic segmentation architecture since it is designed to achieve fast inference. To use DDRNet as encoder, we change the final layer to predict feature maps and then stack three GUBs as shown in Fig.~\ref{fig:GuideDepth}. We down-sample the input image $\mathbf{x}$ for guidance in the first two GUBs, while the last one receives the input image at it's original resolution. Furthermore, the last GUB directly reconstructs the depth map instead of another feature representation. We refer to this architecture as GuideDepth.

\subsection{Network Training}
We follow the training procedure proposed by Alhashim and Wonka \cite{alhashim_high_2018}: Let $\mathbf{\Hat{y}} = f_\theta(\mathbf{x})$ be a depth prediction of the model derived from an input RGB image $\mathbf{x}$ and $\mathbf{y}$ a groundtruth depth map. Then, the objective function is given by:
\begin{equation}
    \mathcal{L}(\mathbf{y}, \hat{\mathbf{y}}) = \mathcal{L}_{DSSIM}(\mathbf{y}, \hat{\mathbf{y}}) + \mathcal{L}_{grad}(\mathbf{y}, \hat{\mathbf{y}}) + \lambda \mathcal{L}_{1}(\mathbf{y}, \mathbf{\Hat{y}})\:
\end{equation}
where the structural dissimilarity loss $\mathcal{L}_{DSSIM}(\mathbf{y}, \mathbf{\Hat{y}}) = \frac{1 - SSIM(\mathbf{y}, \mathbf{\Hat{y}})}{2}$ enforces the model to predict depth maps that are perceived similarly to the groundtruth by using the inverse of the Sturctural Similarity (SSIM) \cite{wang_image_2004}. The gradient loss $\mathcal{L}_{grad} = \frac{1}{P} (\lVert \mathbf{g}_x - \mathbf{\Hat{g}}_x \rVert + \lVert \mathbf{g}_y - \mathbf{\Hat{g}}_y\rVert$) helps the model to learn edges by comparing the pixel-wise partial derivatives $\mathbf{g}_x$ and $\mathbf{g}_y$ of the ground truth in $x$ and $y$ direction with the partial derivative $\mathbf{\Hat{g}}_x$ and $\mathbf{\Hat{g}}_y$ of the prediction. $\mathcal{L}_{1} = \frac{1}{P} \lVert \mathbf{y}- \mathbf{\Hat{y}} \rVert$ increases the overall accuracy of the model through pixel-wise supervision. Hereby $P$ denotes the total number of pixels in a depth map. Moreover, the weighting term $\lambda = 0.1$ balances out the magnitude of the $\mathcal{L}_1$ term as proposed by Alhashim and Wonka \cite{alhashim_high_2018}. To optimize for the objective function, we rely on backpropagation and gradient descent.

\begin{table*}[ht]
\centering
\vspace*{2mm} 
\caption{Model performance on NYU Depth V2 \cite{silberman_indoor_2012}, following the evaluation procedure described in Section \protect\ref{sec:evaluation}. Our models GuideDepth and GuideDepth-S outperform state-of-the-art related methods constantly wrt. accuracy, while having reasonably high throughput for real-time applications. (\textit{$\dagger$: retrained and evaluated with our procedure.})}
\begin{tabular}{l|c|cc|cll|ccc|cc}
\multicolumn{1}{c|}{\multirow{2}{*}{Method}} &
  \multirow{2}{*}{Resolution} &
  \multirow{2}{*}{\begin{tabular}[c]{@{}c@{}}Parameters\\ {[}M{]}\end{tabular}} &
  \multirow{2}{*}{\begin{tabular}[c]{@{}c@{}}MACs\\ {[}G{]}\end{tabular}} &
  \multirow{2}{*}{RMSE} &
  \multirow{2}{*}{rel} &
  \multirow{2}{*}{log10} &
  \multirow{2}{*}{$\delta_1$} &
  \multirow{2}{*}{$\delta_2$} &
  \multirow{2}{*}{$\delta_3$} &
  \multicolumn{2}{c}{fps} \\ 
\multicolumn{1}{c|}{} &         &     &      &       &       &       &       &       &       & Nano   & NX    \\ \hline
FastDepth \cite{wofk_fastdepth_2019}            & 224 $\times$ 224 & 1.3 & 0.37 & 0.604 & -    & -    & 0.771 & -     & -   & 58.8 & -     \\
TuMDE \cite{tu_efficient_2021}             & 224 $\times$ 224 & 1.6 & 0.47  & 0.597  & -    & -    & 0.756 & -    & -    & 83.3 & -      \\

TuMDE \cite{tu_efficient_2021}            & 228 $\times$ 304 & 1.6 & 0.47 & 0.568 & -    & -     & 0.776 & -     & -  & 47.6 &-   \\
\hline
TuMDE $\dagger$ \cite{tu_efficient_2021} &240 $\times$ 320 &5.7 &1.42 &0.531 &0.147 &0.062 &0.801 &0.956 &0.989 &15.5 &- \\
FastDepth $\dagger$ \cite{wofk_fastdepth_2019} & 240  $\times$ 320 & 3.9 & 1.20 & 0.576 & 0.165 & 0.067 & 0.777 & 0.949 & 0.987 & 21.3    & \textbf{149.3}       \\
GuideDepth (Ours)               & 240 $\times$ 320 & 5.8 & 2.63 & \textbf{0.501} & \textbf{0.138} & \textbf{0.058} & \textbf{0.823} & \textbf{0.961} & \textbf{0.990} & \underline{22.9}     & 99.6    \\
GuideDepth-S (Ours)           & 240 $\times$ 320 & 5.7 & 1.52  & \underline{0.514} &\underline{0.144}  & \underline{0.060} & \underline{0.812} & \underline{0.958} & \underline{0.989} & \textbf{35.1}    & \underline{144.5}  \\
\hline
GuideDepth (Ours)               & 480 $\times$ 640 & 5.8 & 10.47 & 0.478 & 0.128 & 0.055 & 0.840 & 0.969 & 0.992 & 6.1   & 29.5    \\
GuideDepth-S (Ours)           & 480 $\times$ 640 & 5.7 & 6.03  & 0.491 & 0.131 & 0.057 & 0.830 & 0.967 & 0.991 & 9.8    & 49.7  
\end{tabular}

\label{table:NYU_RealTime}
\end{table*}
\begin{table*}
\centering
\caption{Model performance on KITTI \cite{geiger_vision_2013}, following the evaluation procedure described in Section \protect\ref{sec:evaluation}. Our models GuideDepth and GuideDepth-S outperform related architectures wrt. accuracy, while having reasonably high throughput for real-time applications. Note, that for resolution 384 \(\times\) 1280, the performance is only reported for the Xavier NX since the Nano can not handle such resolution. (\textit{$\dagger$: retrained and evaluated with our procedure.})}
\begin{tabular}{l|c|cc|cll|ccc|cc}
\multicolumn{1}{c|}{\multirow{2}{*}{Method}} &
  \multirow{2}{*}{Resolution} &
  \multirow{2}{*}{\begin{tabular}[c]{@{}c@{}}Parameters\\ {[}M{]}\end{tabular}} &
  \multirow{2}{*}{\begin{tabular}[c]{@{}c@{}}MACs\\ {[}G{]}\end{tabular}} &
  \multirow{2}{*}{RMSE} &
  \multirow{2}{*}{rel} &
  \multirow{2}{*}{log10} &
  \multirow{2}{*}{$\delta_1$} &
  \multirow{2}{*}{$\delta_2$} &
  \multirow{2}{*}{$\delta_3$} &
  \multicolumn{2}{c}{fps} \\
\multicolumn{1}{c|}{} &          &     &      &       &  &  &       &  &  & Nano & NX \\
\hline
TuMDE $\dagger$ \cite{tu_efficient_2021}            & 192 $\times$ 640  & 5.7  & 2.19      & 5.801  & 0.150 & 0.068 & 0.760   & 0.930 & 0.980   &10.5     &75.7     \\
FastDepth $\dagger$ \cite{wofk_fastdepth_2019}   & 192 $\times$ 640  & 3.9  & 1.82  & 5.839 &0.168  &0.070  & 0.752 &0.927 &0.977  & \underline{15.0} & \underline{101.4}    \\
GuideDepth (Ours)          & 192 $\times$ 640  & 5.8 & 4.19  &\textbf{5.194} &\textbf{0.142}  &\textbf{0.061}  & \textbf{0.799} & \textbf{0.941} & \textbf{0.982}  & 14.9 & 69.3    \\
GuideDepth-S (Ours)           & 192 $\times$ 640  & 5.7 & 2.41  &\underline{5.480} &\underline{0.142}  &\underline{0.063}  & \underline{0.784}  & \underline{0.936} & \underline{0.981}  & \textbf{23.7} & \textbf{102.9}  \\
\hline
GuideDepth (Ours)                & 384 $\times$ 1280 & 5.8  & 16.75     &4.956       &0.133  &0.056  &0.819  &0.952  &0.986  & n/a    & 20.5    \\
GuideDepth-S (Ours)         & 384 $\times$ 1280 & 5.7  & 9.64 & 4.934   &0.133  &0.056  &0.820       &0.952  &0.986  & n/a   & 33.0   
\end{tabular}
\label{table:KITTI_RealTime}
\end{table*}

\section{Evaluation}
\label{sec:evaluation}
We evaluate our approach on two standard benchmarks for monocular depth estimation. Moreover, we perform experiments on different embedded platforms and report results on different ablation studies. In addition to the GuideDepth architecture from Fig.~\ref{fig:GuideDepth}, we also evaluate a smaller model where the number of feature maps in the decoder is half of the original version. We refer to it as \textit{GuideDepth-S}.

\paragraph{Implementation \& Training}
Our method is implemented in PyTorch \cite{paszke_pytorch_2019}. We use the Adam \cite{kingma_adam_2015} optimizer with $\beta_1 = 0.9, \beta_2=0.999$, a learning rate of $0.0001$, and batch size of 8 images. We train for 20 epochs and reduce the learning rate by a factor of 10 after 15 epochs. Additionally, we use data augmentation during training following the same protocol as Alhashim and Wonka~\cite{alhashim_high_2018}, where random horizontal flips ($p=0.5$) and random colour channel swaps ($p=0.25$) are applied. Finally, inverse depth norm is used to bring depth values $y = \frac{d_{max}}{y_{orig}}$ to a predefined range, where $d_max$ is the maximum depth value in the dataset.

\paragraph{Hardware Platforms}
Our proposed architecture targets inference on embedded devices. Hence, we evaluate the model performance on two different NVIDIA Jetson Single Board Computers (SBCs): (1) Jetson Nano and (2) Jetson Xavier NX \footnote{https://developer.nvidia.com/embedded/jetson-modules}. Both boards have similar dimensions but differ significantly in performance. The Jetson Nano employs a quad-core Arm Cortex-A57 processor, a 128-core NVIDIA Maxwell GPU and 4\,GB of RAM. In contrast, the Jetson Xavier NX uses a 6-core NVIDIA Carmel Arm processor, a 384-core NVIDIA Volta GPU and 8\,GB of RAM. The results are reported at 10\,W power mode for the Jetson Nano and 15\,W for the Jetson Xavier NX.

\paragraph{Datasets}
We evaluate on the NYU Depth V2 \cite{silberman_indoor_2012} and the KITTI \cite{geiger_vision_2013} datasets. NYU Depth V2 \cite{silberman_indoor_2012} covers indoor scenes recorded with a Microsoft Kinect camera at a resolution of $480\times640$ pixels with densely annotated ground truth depth values. For training, the reduced dataset proposed by Alhashim and Wonka \cite{alhashim_high_2018} is used, consisting of 50.688 images. The KITTI dataset \cite{geiger_vision_2013} contains 23.158 images from outdoor street scenes with sparse ground-truth depth data, captured by a LIDAR sensor. To provide dense ground-truth depth maps, we rely on the colorization method proposed by Levin \textit{et.~al.}~\cite{levin_colorization_2004}.

\paragraph{Evaluation Metrics}
We make use of standard metrics from the literature~\cite{eigen_depth_2014}. Let $y_i$ be the pixel of ground-truth depth map $\mathbf{y}$ and $\hat{y}_i$ be the pixel in the prediction $\mathbf{\Hat{y}}$, and $P$ be the total number of pixels. Then we rely on the following metrics: Root mean square error $RMSE = \sqrt{\frac{1}{P} \sum^{P}_{i=1} (y_i - \hat{y}_i)^2}$, relative absolute error $rel = \frac{1}{P} \sum^{P}_{i=1} \frac{|y_i - \hat{y}_i|}{y_i}$, scale-invariant error  $log10 = \frac{1}{P} \sum^{P}_{i=1} |\log_{10}(y_i) - \log_{10}(\hat{y}_i)|$ and the threshold accuracy $\delta_j$: $\textnormal{$\%$ of $\hat{i}$ s.t. max} \left( \frac{y_i}{\hat{y}_i}, \frac{\hat{y}_i}{y_i} \right) < 1.25^j, j\in \{1,2,3\}$. 
All results are reported on models converted to tensorRT \cite{nvidia_tensorrt_2021} with quantization to float16. Inference times are averaged over 200 samples.

\paragraph{Evaluation Protocol}
SOTA real-time methods such as FastDepth \cite{wofk_fastdepth_2019} and Tu \textit{et al.} \cite{tu_efficient_2021} evaluate their network performance on center cropped input images to lower the input resolutions of their models. However, this harms comparability as the areas of the prediction differ depending on the crop, effectively only allowing comparison against methods using the exact same crops. We aim at using an evaluation protocol that produces comparable results for models with differing resolutions, reinforcing the need of balancing inference speed and accuracy when reducing resolution.\\
For that purpose, we use the test splits defined by Eigen \textit{et al.} \cite{eigen_depth_2014} for NYU Depth V2 \cite{silberman_indoor_2012} and KITTI \cite{geiger_vision_2013}. For evaluation (and training), the input image is resized to the desired resolution of the model through bilinear interpolation. This allows to produce a low resolution prediction covering the full area of the ground-truth depth map. For the computation of evaluation metrics, the prediction is upsampled to the resolution of the ground-truth depth map. To account for asymmetries, we report the average evaluation result of the test set and the vertically flipped test set as proposed by \cite{alhashim_high_2018}. The test crops by Eigen \cite{eigen_depth_2014} are then used for calculating the evaluation metrics on both predictions at the resolution of the dataset. For NYU Depth V2, the cropped depth map $\mathbf{y}_{20:460, 24:616}$ is considered to exclude noisy ground-truth values at the boarders of the images. For KITTI, the crop excludes area with a low presence of LIDAR points. As the images in KITTI differ slightly in resolution, the crop is recalculated for each image $y \in \mathbb{R}^{H \times W \times 3}$ to evaluate on depth maps of size $\mathbf{y}_{a:b, c:d}$ where $a =0.332H$, $b = 0.914H$, $c = 0.036W$ and $d = 0.964W$.

\subsection{NYU Depth V2 Results}
\begin{figure*}
    \centering
    \subfloat[RGB]{\includegraphics[trim={0.84cm 0.2cm 0.85cm 0cm}, clip, width=.202\textwidth]{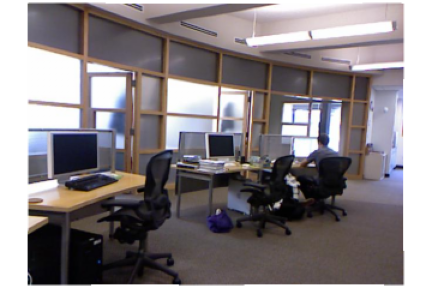}}
    \subfloat[Ground Truth]{\includegraphics[trim={0.8cm 0cm 0.8cm 0cm}, clip, width=.198\textwidth]{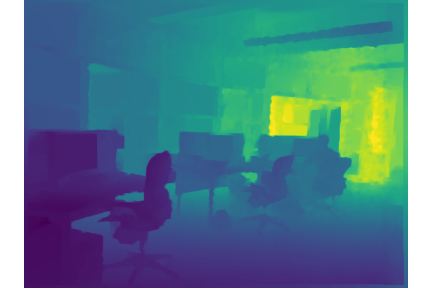}}
    \subfloat[FastDepth $\dagger$ \cite{wofk_fastdepth_2019}]{\includegraphics[trim={0.8cm 0cm 0.8cm 0cm}, clip, width=.198\textwidth]{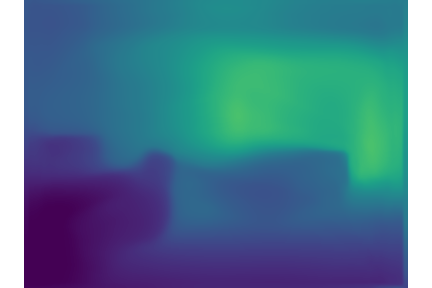}}
    \subfloat[TuMDE $\dagger$ \cite{tu_efficient_2021}]{\includegraphics[trim={0.8cm 0cm 0.8cm 0cm}, clip, width=.198\textwidth]{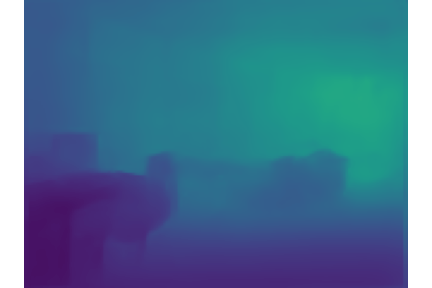}}
    \subfloat[Ours]{\includegraphics[trim={0.8cm 0cm 0.8cm 0cm}, clip, width=.198\textwidth]{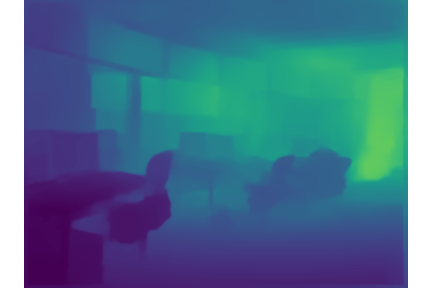}}
    \caption{The qualitative results on NYU Depth V2 \cite{silberman_indoor_2012} show that our predicted depth map is of significantly greater detail. Related methods denoted with $\dagger$ are retrained and evaluated using the procedure described in Section \protect\ref{sec:evaluation}.}
    \label{fig:NYU_VIS}
\end{figure*}

Considering the results in Tab.~\ref{table:NYU_RealTime}, GuideDepth and GuideDepth-S outperform the related work targeting embedded systems in terms of RMSE and $\delta$ accuracy for all experiments. We retrained FastDepth \cite{wofk_fastdepth_2019} and the model proposed by Tu \textit{et al.} \cite{tu_efficient_2021} -- to which we will refer to as TuMDE -- with the same training and evaluation procedure as described in Sec.~\ref{sec:evaluation} (denoted with $\dagger$) to provide direct comparison between their architectures and our model. GuideDepth clearly outperforms the architecture of FastDepth with respect to the accuracy, while achieving comparable throughput. \\
Additionally, the first three columns in Table \ref{table:KITTI_RealTime} showcase the inference speeds reported by Wofk \textit{et al.} \cite{wofk_fastdepth_2019} and Tu \textit{et al.}  \cite{tu_efficient_2021}, indicating a approximate speed-up of factor 2-3 by their compression techniques. However, these models were evaluated on the previously mentioned center-crops, meaning that accuracy and RMSE values are not directly comparable to the further results. In Fig.~\ref{fig:NYU_VIS}, we present a visual comparison of the results.

\begin{figure*}
    \centering
    \subfloat[RGB]{\includegraphics[trim={0.cm 0cm 0cm -0.5cm}, clip, width=.2\textwidth]{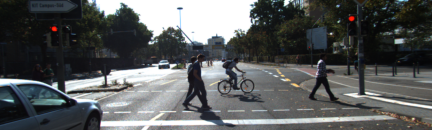}}
    \subfloat[Ground Truth]{\includegraphics[trim={0cm 0cm 0cm 0cm}, clip, width=.2\textwidth]{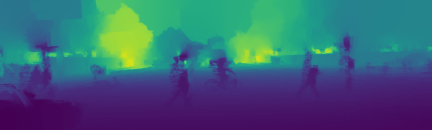}}
    \subfloat[FastDepth $\dagger$ \cite{wofk_fastdepth_2019}]{\includegraphics[trim={0cm 0cm 0.1cm -.2cm}, clip, width=.2\textwidth]{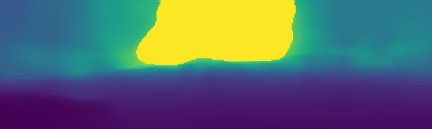}}
    \subfloat[TuMDE $\dagger$ \cite{tu_efficient_2021}]{\includegraphics[trim={0cm 0.08cm 0.1cm -.2cm}, clip, width=.2\textwidth]{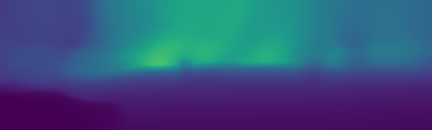}}
    \subfloat[Ours]{\includegraphics[trim={0cm 0cm 0cm 0cm}, clip, width=.2\textwidth]{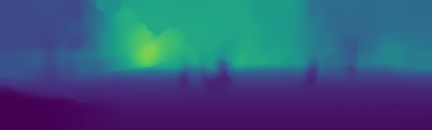}}
    \caption{The qualitative results on KITTI \cite{geiger_vision_2013} show that our predicted depth map is of significantly greater detail, especially when inspecting the pedestrians. For related methods, we show results for retrained models, denoted by $\dagger$.}
    \label{fig:KITTI_VIS}
\end{figure*}

\subsection{KITTI Results}
For the evaluation on KITTI, we retrained TuMDE \cite{tu_efficient_2021} and FastDepth \cite{wofk_fastdepth_2019} with the same training and evaluation procedure as described in Section \ref{sec:evaluation} (denoted with $\dagger$). Note that we did not apply the post-processing steps such as model pruning and only converted models to tensorRT for inference. This results in a decreased throughput compared to the results reported by Wofk \textit{et al.} \cite{wofk_fastdepth_2019} and Tu \textit{et al.} \cite{tu_efficient_2021}. Furthermore, using our evaluation protocol described in Section \ref{sec:evaluation} makes the benchmark more challenging for models operating on low resolution, resulting in higher errors.\\
In this experiment, our proposed method GuideDepth and GuideDepth-S outperform the related methods in terms of RMSE and $\delta_j$ accuracy for all experiments. Additionally, when solely comparing the architectures without considering post-processing, GuideDepth-S outperforms FastDepth on the inference speed. Considering the results on the resolution of 384 \(\times\) 1280, GuideDepth-S surprisingly achieves similar accuracy to the standard model. Note that we only report results on the Jetson Xavier NX for this resolution, since the computational requirements exceed the capabilities of the Jetson Nano. In Fig.~\ref{fig:KITTI_VIS}, we compare the visual results of related architectures with our model, showing the improvements gained by our approach. Especially the model's capabilities to capture pedestrians in the depth prediction stands out when comparing to the results of related architectures.

\subsection{Ablation Studies}
We perform various ablation studies on the NYU Depth V2 \cite{silberman_indoor_2012} dataset based on the previously described evaluation and training procedures, using the image resolution of $240\times320$ pixels. The results are presented in Table \ref{table:Ablation_Guidance} and Table \ref{table:encoders}.

\subsubsection{Guidance}
\label{Guidance}
This experiment investigates the importance of the guidance image and the corresponding processing branch. In particular, we compare the usage of the guidance image in the Guided Upsampling Block as proposed in the method section against directly concatenating the guidance image with the feature maps without the preceding feature extraction branch (Direct guidance in Table \ref{table:Ablation_Guidance}). Inspired by Song \textit{et al.} \cite{song_monocular_2021}, who relied on images from the Laplacian pyramid in their decoder architecture, Laplacian images are also used as guidance in our second experiment. To increase computational efficiency, the Laplacian images $\textbf{L}_k$ are generated by $\mathbf{L}_k=Up_k(Down_{k+1}(\mathbf{x}))$, where $k$ describes the target scale $s=\frac{1}{2^k}$ of the interpolated image for $k\in\{0,1,2\}$. Additionally, a model without a guidance image is included in the comparison.\\
Investigating the results, the usage of Laplacian images does not introduce any benefits in terms of prediction quality. This could be because the GUB learns to extract diverse, low-level features from the rich image while the Laplacian image already contains reduced information in the form of edges. \\
Concatenating the guidance image directly with the feature maps results in faster inference and, at first glance, only shows a minor difference for the accuracy and error when compared to the significantly slower GUB. However, we notice that the visual quality of depth predictions when using GUB is noticeable better, justifying our choice for the usage of the GUB over direct image guidance.

\begin{table}[ht]
\centering
\caption{Ablation study on different types of guidance images (RGB image and Laplacian image) as well as modifications to the GUB architecture. The last row shows results when using no guidance image.}
\begin{tabular}{ll|c|c|ccc}
\multicolumn{2}{c|}{Guidance}                          & \multicolumn{1}{c|}{Inference} & \multicolumn{1}{c|}{\multirow{2}{*}{RMSE}} & \multicolumn{1}{c}{\multirow{2}{*}{$\delta_1$}} & \multicolumn{1}{c}{\multirow{2}{*}{$\delta_2$}} & \multicolumn{1}{c}{\multirow{2}{*}{$\delta_3$}} \\
\multicolumn{1}{c}{Type} & \multicolumn{1}{c|}{Branch} & \multicolumn{1}{c|}{{[}ms{]}}  & \multicolumn{1}{c|}{}                      & \multicolumn{1}{c}{}                            & \multicolumn{1}{c}{}                            & \multicolumn{1}{c}{}                            \\\hline
Image     & GUB    & 44 & 0.501 & 0.823 & 0.961 & 0.990 \\
Image     & Direct & 37 & 0.508 & 0.821 & 0.961 & 0.990 \\
Laplacian & GUB    & 44 & 0.508 & 0.819 & 0.961 & 0.990 \\
Laplacian & Direct & 38 & 0.509 & 0.818 & 0.962 & 0.990 \\
None      & None   & 34 & 0.510 & 0.821 & 0.961 & 0.990
\end{tabular}
\label{table:Ablation_Guidance}
\end{table}

\subsubsection{Encoder}
Since the decoder was designed to be combined with Hong \textit{et al.}'s DDRNet \cite{hong_deep_2021} encoder, we investigate the impact of employing different encoders. General purpose backbones like MobileNetV2 \cite{sandler_mobilenetv2_2018} and HarDNet \cite{chao_hardnet_2019} usually extract features at lower resolution with a higher number of channels. Therefore, we upsample these feature maps based on two stages of the decoder from FastDepth \cite{wofk_fastdepth_2019}, as the information from guidance images of our decoder at this resolution is not useful. Then, our decoder is attached as described in the method section.\\
Furthermore, we can compare with FastDepth \cite{wofk_fastdepth_2019}, as it is based on MobileNetV2 \cite{sandler_mobilenetv2_2018}) for the encoder. By comparing the results in Table \ref{table:encoders}, we note that our decoder improves the RMSE and $\delta_1$ accuracy while marginally impacting throughput. HarDNet-39-DS increases the prediction accuracy even further at similar throughput. The usage of DDRNet-23-slim, finally, allows to reach the desired accuracy while additionally reducing the inference time, being the most powerful encoder for the model. Interestingly, DDRNet-23-slim allows faster inference than the other encoders although having more parameters and MACs. 

\begin{table}[thpb]
\centering
\scriptsize
\caption{Comparison of different encoders combined with our decoder. Using MobileNetV2 \cite{sandler_mobilenetv2_2018} allows direct comparison between our decoder and the retrained FastDepth architecture \cite{wofk_fastdepth_2019}.}
\begin{tabular}{l|cc|c|c|c}
\multicolumn{1}{c|}{Encoder} &
  \begin{tabular}[c]{@{}c@{}}Params.\\ {[}M{]}\end{tabular} &
  \begin{tabular}[c]{@{}c@{}}MACs\\ {[}G{]}\end{tabular} &
  \multicolumn{1}{c|}{\begin{tabular}[c]{@{}c@{}}Inference\\ {[}ms{]}\end{tabular}} &
  \multicolumn{1}{c|}{RMSE} &
  \multicolumn{1}{c}{$\delta_1$} \\
  \hline
FastDepth $\dagger$ \cite{wofk_fastdepth_2019}    & 3.9 & 1.20 & 45 & 0.576 & 0.777 \\
\hline
MobileNetV2 \cite{sandler_mobilenetv2_2018}  & 4.7    & 2.27     & 53 & 0.547 & 0.790 \\
HarDNet39-DS \cite{chao_hardnet_2019}  & 2.9    & 2.09     & 52 & 0.516 & 0.811 \\
DDRNet-23-slim \cite{hong_deep_2021} & 5.8 & 2.63 & 44 & 0.501 & 0.823
\end{tabular}
\label{table:encoders}
\end{table}

\section{CONCLUSIONS}

We presented an encoder-decoder architecture for monocular depth estimation, targeted to robotics systems with constrained resources. Our approach focused on accelerating the decoding part while relying on a standard encoder. We proposed the Guided Upsampling Block (GUB) for guiding the decoder on upsampling the feature representation and the final depth map, allowing us to achieve more detailed, high-resolution depth predictions. By stacking multiple GUBs together, we designed a cost-efficient decoder that has a good balance between accuracy and speed when compared with related architectures on KITTI and NYU Depth V2 datasets. For future work, we aim to examine our models capabilities when training directly on the embedded platform, similar to domain adaption on resource-constrained hardware~\cite{hornauer_visual_2021}.

%%%%%%%%%%%%%%%%%%%%%%%%%%%%%%%%%%%%%%%%%%%%%%%%%%%%%%%%%%%%%%%%%%%%%%%%%%%%%%%%
\section*{ACKNOWLEDGMENT}
The authors acknowledge support by the state of Baden-Württemberg through bwHPC.

%%%%%%%%%%%%%%%%%%%%%%%%%%%%%%%%%%%%%%%%%%%%%%%%%%%%%%%%%%%%%%%%%%%%%%%%%%%%%%%%
\bibliographystyle{ieeetran}
\bibliography{references}

\end{document}